\documentclass[conference]{IEEEtran}
\IEEEoverridecommandlockouts
\usepackage{cite}
\usepackage{amsmath,amssymb,amsfonts}
\usepackage{algorithmic}
\usepackage{graphicx}
\usepackage{textcomp}
\usepackage[numbib]{tocbibind}
\usepackage[svgnames]{xcolor}
\usepackage{xcolor}
\usepackage[colorlinks = true,
            linkcolor = blue,
            urlcolor  = blue,
            citecolor = blue,
            anchorcolor = blue]{hyperref}

\newcommand{\MYhref}[3][blue]{\href{#2}{\color{#1}{#3}}}%

\def\BibTeX{{\rm B\kern-.05em{\sc i\kern-.025em b}\kern-.08em
    T\kern-.1667em\lower.7ex\hbox{E}\kern-.125emX}}

\begin{document}

\title{Is Attentional Channel Processing Design Required?\\Comprehensive Analysis Of Robustness Between Vision Transformers And Fully Attentional Networks \\
}

\author{\IEEEauthorblockN{Abhishri Ajit Medewar}
\IEEEauthorblockA{\textit{Computer Science, SCAI} \\
\textit{Arizona State University}\\
Tempe, AZ, USA \\
amedewar@asu.edu}
\and
\IEEEauthorblockN{Swanand Ashokrao Kavitkar}
\IEEEauthorblockA{\textit{Computer Science, SCAI} \\
\textit{Arizona State University}\\
Tempe, AZ, USA \\
skavitka@asu.edu}
}

\maketitle

\begin{abstract}
The robustness testing has been performed for standard CNN models \cite{IRCOA} and Vision Transformers, however there is a lack of comprehensive study between the robustness of traditional Vision Transformers without an extra attentional channel design and the latest fully attentional network(FAN) models. So in this paper, we use the ImageNet dataset to compare the robustness of fully attentional network(FAN) models with traditional Vision Transformers to understand the role of an attentional channel processing design using white box attacks and also study the transferability between the same using black box attacks. 
\\Project page : \MYhref{https://github.com/S-Threepio/Robustness-Analysis-FAN-ViT}{\underline {GITHUB LINK}}
\end{abstract}

\section{\textbf{Problem Definition}}
Image classification is one of the crucial component of computer vision. The performance of many modern age applications such as Autonomous driving, Anomaly detection, Inventory management, Object detection is centered around the accuracy of the underlying classifier. Accuracy is an important metric for the performance of such models, however robustness of such models is often penalized to achieve better prediction results \cite{IRCOA}. Along with CNN now there has been an increase in adaption of Vision transformers in object detection, segmentation, image classification, and action recognition. Our study is focused on the FAN ViT models which are proven to be more robust than the traditional ViT. We propose a comprehensive study on the comparison of robustness between FAN ViTs and Traditional ViT using the standard evaluation metrics to determine the importance of attentional channel processing design .

\section{\textbf{Challenges and Main Contributions}}
The core challenges of this study are as below :
\\Understanding the implementation of the FAN ViT models with attentional channel process design. Finding how the differences between FAN ViTs and Traditional ViT  affect the robustness of the models.
\\Understanding the mathematical reasoning behind the difference of robustness between FAN ViTs and traditional ViT
\\Our main contribution in this paper is to give a comprehensive analysis on the robustness of FAN ViTs compared to Traditional Vit. 
\section{\textbf{Most related prior work and its shortcomings}}
\subsection{\textbf{Is Robustness the Cost of Accuracy?}
– A Comprehensive Study on the Robustness of
18 Deep Image Classification Models}
\cite{IRCOA} This paper focuses on the robustness of the then robust CNN architecture using the standard evaluation metrics and below approaches to generate adversarial attacks using white box attacks:
a) FGSM b) Iterative FGSM c) C\&W attack d) EAD-L1 attack
\textbf{Although it covers the mentioned white box attacks,it lacks the detailed robustness evaluation using black box attacks.
}
\subsection{\textbf{Understanding The Robustness in Vision Transformers}}
\cite{UTRVT} This paper focuses on the robustness comparison between FAN ViT models and the State of the art CNN models. The key focus of the paper is to propose a new architecture called FAN ViT and it's variations to increase robustness in traditional ViT. \textbf{However, the paper lacks the comprehensive study on the robustness comparison between traditional ViT models and FAN ViT models.} We propose a robustness comparison focused approach with various attack approaches and standard evaluation metrics to ascertain the need of attentional channel processing design.
\subsection{\textbf{On the Robustness of Vision Transformers to Adversarial Examples}
}
\cite{OTROVTTAE} This paper focuses on the robustness of vision transformers to adversarial attacks.
It also proposes a new method of robustness testing called self-attention blended gradient attack to show the vulnerability of model to white box attacks
The study encompasses multiple Vision Transformers, Big Transfer Models and CNN architectures trained on CIFAR-10, CIFAR-100 and ImageNet.
\textbf{However the paper lacks the comparison of the latest FAN ViT models
}
\subsection{\textbf{Patch-Fool: Are Vision Transformers Always Robust Against Adversarial Perturbation?}} 
\cite{PF} This paper focuses on a new approach to study the robustness of vision transformers called patch fool which fools the self-attention mechanism by
attacking its basic component (i.e., a single patch) with a series of attention-aware
optimization techniques. \textbf{Although this study gives a detailed perspective on the robustness testing of ViTs, it revolves around the Patch fool method and lacks an analysis on different attack approaches.}

\subsection{\textbf{Understanding Robustness of Transformers for Image Classification}
}
\cite{URTFIT} This study produces a variety of different measures of robustness of ViT models
and compare the findings to ResNet baselines. \textbf{But the latest FAN models developed with attentional channel processing design are not explored in this paper.}

\section{\textbf{Methodology}}
We will be using the existing models of the latest FAN ViTs and traditional ViTs to generate adversarial examples and robustness testing.
The traditional ViT model will be our baseline to compare the robustness of FAN ViT models. We will be using DeiT-S- Data Efficient Image Transformer-Small.
We are going to use 6 different FAN ViT models as below :
\begin{enumerate}
\item FAN-tiny-ViT
\item FAN-small-ViT
\item FAN-base-ViT
\item FAN-tiny-Hybrid
\item FAN-small-Hybrid
\item FAN-base-Hybrid
\end{enumerate}
The core difference between \cite{UTRVT} FAN-ViTs and FAN-Hybrid models is that Hybrid models will have convolution blocks for bottom two stages with down sampling. Each stage has 3 convolution blocks.
Refer the paper \cite{UTRVT} for detailed understanding of each FAN model.
We will use below 4 attacks to generate adversarial images for each model and to analyse the robustness.

\begin{enumerate}
\item Fast Gradient Sign Method(FGSM)
\item Projected Gradient Descent(PGD)
\item Iterative FGSM
\item Momentum Iterative Method(MiM)
\end{enumerate}

\section{\textbf{Data to be used}}
\begin{figure}[ht]
\includegraphics[height=5cm,width=5cm]{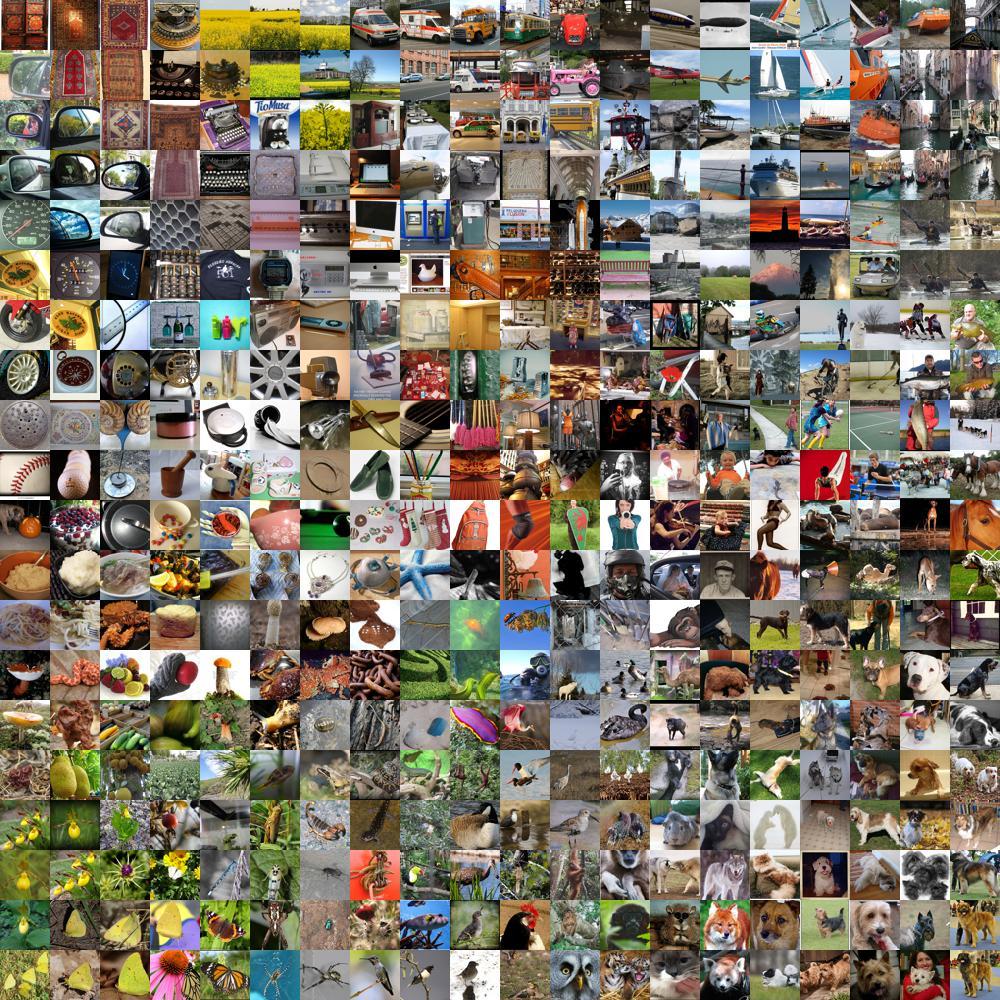}
\centering
\caption{ImageNet dataset}
\end{figure}
We will be using the ImageNet dataset for our purpose of analysis since most of the CNN models and ViT were designed for Imagenet challenge. The ImageNet dataset has a huge dataset comprising of millions of images. The dataset is divided into training, validation and testing set. Samples from testing set will be used to generate adversarial examples for robustness analysis. 
\\For calculating the model accuracy we will use the full dataset of 50000 images. However for generating the adversarial examples we will be using randomly sampled 1000 images and give an estimate of robustness based on that. 

\section{\textbf{Experimental Setup}}
As mentioned in our approach we first ran the Imagenet dataset with 50000 images and calculated the accuracy for our baseline traditional ViT model as well as various FAN Models. Following that we created adversarial images using above mentioned attacks. For creating adversarial images we have run the attacks for 1000 images sampled randomly from Imagenet validation dataset. We then calculated the attack accuracy for the traditional ViT model as well as FAN models using these perturbed images.
\\After attacking each model with the proposed methods we will then calculate the impact on each of the model using below evaluation metrics :
\\1) \textbf{Attack success rate :} 
\\We will first check if the model was able to predict the correct label for a given image. If the model predicted the correct label only then we will perturb the image data and test it again. Reason for this is perturbing an image which was incorrectly predicted before the attack will not be an accurate evaluation of the attack accuracy.
Further we will calculate the number of images that were incorrectly predicted after the attack as compared to the images that were correctly predicted before the attack.
So our Attack success rate will be :
\begin{displaymath}
Attack Accuracy = 
\frac{incorrectly\ predicted\ labels\ after\ attack}{correctly\ predicted\ labels\ before\ attack}
\end{displaymath}
2) \textbf{$L_{2}$ and $L_{\infty}$  distortion metric scale :}
\\We use the $L_{2}$ and $L_{\infty}$ to measure the similarity between the original image and the adversarial image. $L_{2}$ distance is basically the absolute difference between each pixel value whereas $L_{\infty}$ is the maximum difference between any two corresponding pixels of the adversarial and original image. These metrics will allow us to understand amount of distortion on the adversarial image compared to original image. It will also be a good estimate on whether the image is able to fool a human being.
\\3) \textbf{Transferability between FANs and traditional ViTs :}   
Considering an attacker does not have access to FAN models, we need to understand the vulnerability of FAN models to black box attacks. For this purpose we will be using 100 adversarial images which are correctly predicted on the traditional ViT model, Then we will perturb the image data and check if it is incorrectly predicted by the traditional model. Such a perturbed image will then be tested on the FAN models. 
We will give the transferability metric as below :
\begin{displaymath}
Transferability = 
\frac{adv\ imgs\ working\ on\ FAN\ vits}{adv\ imgs\ working\ on\ TRAD\ vits}
\end{displaymath}

Based on the above evaluation metrics we will provide a detailed comprehensive analysis on the robustness comparison of FAN ViTs and Traditional ViT.

\section{\textbf{Results}}
As mentioned in our approach we first ran the Imagenet dataset with 50000 images and calculated the model accuracy for our baseline traditional ViT model as well as FAN Models. We got the attack accuracies as well as L2 and Linf distances for the above mentioned attacks for 1000 images. Additionally to test black box attack we attacked all the FAN ViT models using 100 adversarial images developed on traditional ViTs. 
\\The tables II to XII demonstrates the results that were obtained.
It is evident that the FAN Base ViT model perform better amongst all the FAN models. FAN-tiny-ViT and FAN-tiny-hybrid ViTs perform worse than the traditional ViT due to less number of parameters.
\begin{table}
\centering
\caption{}
\begin{tabular}{ |p{2.25cm}||p{2.25cm}|}
 \hline
 \multicolumn{2}{|c|}{Model Accuracies} \\
 \hline
 Model name& Accuracy \\
 \hline
 Baseline DeiT-S& \textbf{81.574}\\
 FAN-tiny-ViT&\textbf{79.108}\\
 FAN-small-ViT&\textbf{82.586}\\
 FAN-base-ViT&\textbf{83.504}\\
 FAN-tiny-Hybrid&\textbf{80.024}\\
 FAN-small-Hybrid&\textbf{83.514}\\
 FAN-base-Hybrid&\textbf{83.810}\\
 \hline
 \end{tabular}
\end{table}

\begin{table}
\centering
\caption{\label{FGSM} }
\begin{tabular}{ |p{2.15cm}||p{1.5cm}||p{1.5cm}||p{1.5cm}|}
\hline
\multicolumn{4}{|c|}{FGSM Attack} \\
\hline
 Model name& Attack Accuracy& L2 distance& Linf distance \\
 \hline
 Baseline DeiT-S& \textbf{38.27}& \textbf{0.5933}& \textbf{2.013}\\
 FAN-tiny-ViT&\textbf{46.71}& \textbf{0.5887}& \textbf{2.021}\\
 FAN-small-ViT&\textbf{36.88}& \textbf{0.5933}& \textbf{2.017}\\
 FAN-base-ViT&\textbf{31.94}& \textbf{0.590}& \textbf{2.020}\\
 FAN-tiny-Hybrid&\textbf{45.07}& \textbf{0.5916}& \textbf{2.024}\\
 FAN-small-Hybrid&\textbf{34.28}& \textbf{0.585}& \textbf{2.021}\\
 FAN-base-Hybrid&\textbf{36.23}& \textbf{0.577}& \textbf{2.013}\\
 \hline
\end{tabular}
\end{table}

\begin{table}
\centering
\caption{}
\begin{tabular}{ |p{2.15cm}||p{1.5cm}||p{1.5cm}||p{1.5cm}|}
\hline
\multicolumn{4}{|c|}{MiM Attack} \\
\hline
 Model name& Attack Accuracy & L2 distance& Linf distance \\
 \hline
 Baseline DeiT-S& \textbf{91.59}& \textbf{0.572}& \textbf{2.017}\\
 FAN-tiny-ViT&\textbf{93.50}& \textbf{0.584}& \textbf{2.025}\\
 FAN-small-ViT&\textbf{91.24}& \textbf{0.585}& \textbf{2.019}\\
 FAN-base-ViT&\textbf{81.25}& \textbf{0.581}& \textbf{2.023}\\
 FAN-tiny-Hybrid&\textbf{93.24}& \textbf{0.579}& \textbf{2.019}\\
 FAN-small-Hybrid&\textbf{91.67}& \textbf{0.586}& \textbf{2.022}\\
 FAN-base-Hybrid&\textbf{92.38}& \textbf{0.5931}& \textbf{2.023}\\
 \hline
\end{tabular}
\end{table}

\begin{table}
\centering
\caption{}
\begin{tabular}{ |p{2.15cm}||p{1cm}||p{1cm}||p{1cm}||p{1cm}|}
\hline
\multicolumn{5}{|c|}{PGD Attack} \\
\hline
 Model name& Step 1 & Step 2& Step 5 & Step 10 \\
 \hline
 Baseline DeiT-S& \textbf{18.23}& \textbf{45.25}& \textbf{85.14}& \textbf{96.79}\\
 FAN-tiny-ViT&\textbf{29.72}& \textbf{58.24}& \textbf{83.59}& \textbf{94.82}\\
 FAN-small-ViT& \textbf{24.59}& \textbf{44.14}& \textbf{81.72}& \textbf{94.32}\\
 FAN-base-ViT& \textbf{17.26}& \textbf{40.61}& \textbf{69.36}& \textbf{89.36}\\
 FAN-tiny-Hybrid& \textbf{26.54}& \textbf{58.50}& \textbf{87.63}& \textbf{98.10}\\
 FAN-small-Hybrid& \textbf{19.47}& \textbf{44.84}& \textbf{80.43}& \textbf{94.73}\\
 FAN-base-Hybrid& \textbf{21.77}& \textbf{43.91}& \textbf{79.30}& \textbf{94.65}\\
 \hline
\end{tabular}
\end{table}

\begin{table}
\centering
\caption{}
\begin{tabular}{ |p{2.15cm}||p{1cm}||p{1cm}||p{1cm}||p{1cm}|}
\hline
\multicolumn{5}{|c|}{PGD Attack L2 distance} \\
\hline
 Model name& Step 1 & Step 2& Step 5 & Step 10 \\
 \hline
 Baseline DeiT-S& \textbf{0.575}& \textbf{0.573}& \textbf{0.583}& \textbf{0.584}\\
 FAN-tiny-ViT&\textbf{0.590}& \textbf{0.577}& \textbf{0.582}& \textbf{0.580}\\
 FAN-small-ViT& \textbf{0.585}& \textbf{0.594}& \textbf{0.579}& \textbf{0.585}\\
 FAN-base-ViT& \textbf{0.589}& \textbf{0.583}& \textbf{0.583}& \textbf{0.579}\\
 FAN-tiny-Hybrid& \textbf{0.590}& \textbf{0.583}& \textbf{0.580}& \textbf{0.584}\\
 FAN-small-Hybrid& \textbf{0.596}& \textbf{0.579}& \textbf{0.581}& \textbf{0.583}\\
 FAN-base-Hybrid& \textbf{0.592}& \textbf{0.5785}& \textbf{0.5788}& \textbf{0.5768}\\
 \hline
\end{tabular}
\end{table}

\begin{table}
\centering
\caption{}
\begin{tabular}{ |p{2.15cm}||p{1cm}||p{1cm}||p{1cm}||p{1cm}|}
\hline
\multicolumn{5}{|c|}{PGD Attack Linf distance} \\
\hline
 Model name& Step 1 & Step 2& Step 5 & Step 10 \\
 \hline
 Baseline DeiT-S& \textbf{2.010}& \textbf{2.016}& \textbf{2.015}& \textbf{2.025}\\
 FAN-tiny-ViT&\textbf{2.024}& \textbf{2.016}& \textbf{2.017}& \textbf{2.025}\\
 FAN-small-ViT& \textbf{2.018}& \textbf{2.022}& \textbf{2.015}& \textbf{2.018}\\
 FAN-base-ViT& \textbf{2.020}& \textbf{2.022}& \textbf{2.025}& \textbf{2.015}\\
 FAN-tiny-Hybrid& \textbf{2.023}& \textbf{2.016}& \textbf{2.025}& \textbf{2.023}\\
 FAN-small-Hybrid& \textbf{2.020}& \textbf{2.018}& \textbf{2.014}& \textbf{2.017}\\
 FAN-base-Hybrid& \textbf{2.013}& \textbf{2.023}& \textbf{2.024}& \textbf{2.028}\\
 \hline
\end{tabular}
\end{table}

\begin{table}
\centering
\caption{}
\begin{tabular}{ |p{2.15cm}||p{1cm}||p{1cm}||p{1cm}||p{1cm}|}
\hline
\multicolumn{5}{|c|}{BIM Attack} \\
\hline
 Model name& Step 1 & Step 2& Step 5 & Step 10 \\
 \hline
 Traditional ViT& \textbf{11.12}& \textbf{25.86}& \textbf{55.35}& \textbf{76.78}\\
 FAN-tiny-ViT&\textbf{17.75}& \textbf{41.17}& \textbf{66.99}& \textbf{84.082}\\
 FAN-small-ViT& \textbf{14.47}& \textbf{33.21}& \textbf{64.58}& \textbf{81.37}\\
 FAN-base-ViT& \textbf{13.30}& \textbf{27.64}& \textbf{54.32}& \textbf{72.58}\\
 FAN-tiny-Hybrid& \textbf{15.84}& \textbf{36.28}& \textbf{69.06}& \textbf{88.79}\\
 FAN-small-Hybrid& \textbf{14.35}& \textbf{31.67}& \textbf{62.33}& \textbf{81.93}\\
 FAN-base-Hybrid& \textbf{11.59}& \textbf{28.84}& \textbf{62.30}& \textbf{81.44}\\
 \hline
\end{tabular}
\end{table}

\begin{table}
\centering
\caption{}
\begin{tabular}{ |p{2.15cm}||p{1cm}||p{1cm}||p{1cm}||p{1cm}|}
\hline
\multicolumn{5}{|c|}{BIM Attack L2 distance} \\
\hline
 Model name& Step 1 & Step 2& Step 5 & Step 10 \\
 \hline
 Baseline DeiT-S& \textbf{0.086}& \textbf{0.082}& \textbf{0.085}& \textbf{0.087}\\
 FAN-tiny-ViT&\textbf{0.084}& \textbf{0.0861}& \textbf{0.088}& \textbf{0.0861}\\
 FAN-small-ViT& \textbf{0.0874}& \textbf{0.0876}& \textbf{0.08668}& \textbf{0.08662}\\
 FAN-base-ViT& \textbf{0.0825}& \textbf{0.0852}& \textbf{0.0834}& \textbf{0.0887}\\
 FAN-tiny-Hybrid& \textbf{0.085}& \textbf{0.0836}& \textbf{0.0864}& \textbf{0.078}\\
 FAN-small-Hybrid& \textbf{0.0825}& \textbf{0.0858}& \textbf{0.0821}& \textbf{0.0885}\\
 FAN-base-Hybrid& \textbf{0.0856}& \textbf{0.0836}& \textbf{0.0845}& \textbf{0.0863}\\
 \hline
\end{tabular}
\end{table}

\begin{table}
\centering
\caption{}
\begin{tabular}{ |p{2.15cm}||p{1cm}||p{1cm}||p{1cm}||p{1cm}|}
\hline
\multicolumn{5}{|c|}{BIM Attack Linf distance} \\
\hline
 Model name& Step 1 & Step 2& Step 5 & Step 10 \\
 \hline
 Baseline DeiT-S& \textbf{1.496}& \textbf{1.476}& \textbf{1.4780}& \textbf{1.4786}\\
 FAN-tiny-ViT&\textbf{1.485}& \textbf{1.488}& \textbf{1.485}& \textbf{1.481}\\
 FAN-small-ViT& \textbf{1.489}& \textbf{1.499}& \textbf{1.476}& \textbf{1.4819}\\
 FAN-base-ViT& \textbf{1.487}& \textbf{1.487}& \textbf{1.485}& \textbf{1.4874}\\
 FAN-tiny-Hybrid& \textbf{1.49750}& \textbf{1.49795}& \textbf{1.4790}& \textbf{1.461}\\
 FAN-small-Hybrid& \textbf{1.474}& \textbf{1.487}& \textbf{1.473}& \textbf{1.491}\\
 FAN-base-Hybrid& \textbf{1.479}& \textbf{1.491}& \textbf{1.484}& \textbf{1.492}\\
 \hline
\end{tabular}

\end{table}

\begin{table}
\centering
\caption{}
\begin{tabular}{ |p{2.15cm}||p{2.25cm}||p{2.50cm}|}
\hline
\multicolumn{3}{|c|}{Transferability} \\
\hline
 Model name& FGSM Transferability& MiM Transferability \\
 \hline
 FAN-tiny-ViT&\textbf{0.64}& \textbf{1.0}\\
 FAN-small-ViT&\textbf{0.56}& \textbf{1.0}\\
 FAN-base-ViT&\textbf{0.51}& \textbf{1.0}\\
 FAN-tiny-Hybrid&\textbf{0.59}& \textbf{1.0}\\
 FAN-small-Hybrid&\textbf{0.56}& \textbf{1.0}\\
 FAN-base-Hybrid&\textbf{0.53}& \textbf{1.0}\\
 \hline
\end{tabular}
\end{table}

\begin{table}
\centering
\caption{}
\begin{tabular}{ |p{2.15cm}||p{1cm}||p{1cm}||p{1cm}||p{1cm}|}
\hline
\multicolumn{5}{|c|}{PGD Attack Transferability} \\
\hline
 Model name& Step 1 & Step 2& Step 5 & Step 10 \\
 \hline
 FAN-tiny-ViT&\textbf{0.73}& \textbf{0.53}& \textbf{0.42}& \textbf{0.40}\\
 FAN-small-ViT& \textbf{0.55}& \textbf{0.47}& \textbf{0.33}& \textbf{0.26}\\
 FAN-base-ViT& \textbf{0.41}& \textbf{0.40}& \textbf{0.28}& \textbf{0.27}\\
 FAN-tiny-Hybrid& \textbf{0.67}& \textbf{0.51}& \textbf{0.29}& \textbf{0.40}\\
 FAN-small-Hybrid& \textbf{0.54}& \textbf{0.38}& \textbf{0.29}& \textbf{0.25}\\
 FAN-base-Hybrid& \textbf{0.48}& \textbf{0.42}& \textbf{0.24}& \textbf{0.30}\\
 \hline
\end{tabular}
\end{table}

\begin{table}
\centering
\caption{}
\begin{tabular}{ |p{2.15cm}||p{1cm}||p{1cm}||p{1cm}||p{1cm}|}
\hline
\multicolumn{5}{|c|}{BIM Attack Transferability} \\
\hline
 Model name& Step 1 & Step 2& Step 5 & Step 10 \\
 \hline
 FAN-tiny-ViT&\textbf{0.85}& \textbf{0.93}& \textbf{1.0}& \textbf{1.0}\\
 FAN-small-ViT& \textbf{0.76}& \textbf{0.94}& \textbf{1.0}& \textbf{1.0}\\
 FAN-base-ViT& \textbf{0.65}& \textbf{0.88}& \textbf{1.0}& \textbf{1.0}\\
 FAN-tiny-Hybrid& \textbf{0.75}& \textbf{0.96}& \textbf{0.99}& \textbf{1.0}\\
 FAN-small-Hybrid& \textbf{0.67}& \textbf{0.91}& \textbf{0.99}& \textbf{0.99}\\
 FAN-base-Hybrid& \textbf{0.70}& \textbf{0.90}& \textbf{1.0}& \textbf{1.0}\\
 \hline
\end{tabular}
\end{table}

\section{\textbf{Conclusion}}
After performing a detailed analysis on the obtained results we can assert that, the presence of attentional Channel Processing Design in FAN ViT models improve the performance of the traditional ViT. As per the analysis on white box attack accuracies, \textbf{we can conclude that FAN ViTs are less susceptible to adversarial attacks as compared to traditional Vit.}
\\Secondly the L2 and Linf distances for traditional ViT and FAN ViTs are almost similar. This validates the fact that amount of perturbations applied for all the models are of the same concentration.
\\To add to that we tested the transferability of the model by using black box attack. The percentage of adversarial images that worked on FAN ViTs were less than those which worked on traditional ViT. 
\textbf{This result asserts that ViT models with additional attentional channel layer are less susceptible to black box attack.}
\bibliographystyle{unsrt}
\bibliography{references}
\vspace{12pt}

\end{document}